\documentclass{article}

\usepackage[preprint]{neurips_2025}

\usepackage[utf8]{inputenc} 
\usepackage[T1]{fontenc}    
\usepackage{hyperref}       
\usepackage{url}            
\usepackage{booktabs}       
\usepackage{amsfonts}       
\usepackage{nicefrac}       
\usepackage{microtype}      
\usepackage{xcolor}         
\usepackage{graphicx}
\usepackage{amsmath}
\usepackage{amssymb}

\title{Fourier Asymmetric Attention on Domain Generalization for Pan-Cancer Drug Response Prediction}

\author{
Ran Song, Yinpu Bai and Hui Liu$^{*}$ \\
College of Computer and Information Engineering, 
Nanjing Tech University,\\ Nanjing, 211800, China.\\
  \texttt{hliu@njtech.edu.cn}\\
}

\begin{document}

\maketitle

\begin{abstract}
The accurate prediction of drug responses remains a formidable challenge, particularly at the single-cell level and in clinical treatment contexts. Some studies employ transfer learning techniques to predict drug responses in individual cells and patients, but they require access to target-domain data during training, which is often unavailable or only obtainable in future. In this study, we propose a novel domain generalization framework, termed FourierDrug, to address this challenge. Given the extracted feature from expression profile, we performed Fourier transforms and then introduced an asymmetric attention constraint that would cluster drug-sensitive samples into a compact group while drives resistant samples dispersed in the frequency domain. Our empirical experiments demonstrate that our model effectively learns task-relevant features from diverse source domains, and achieves accurate predictions of drug response for unseen cancer type. When evaluated on single-cell and patient-level drug response prediction tasks, FourierDrug—trained solely on \textit{in vitro} cell line data without access to target-domain data—consistently outperforms or, at least, matched the performance of current state-of-the-art methods. These findings underscore the potential of our method for real-world clinical applications. The source code and datasets for reproducing our experiments are available at GitHub (\url{https://anonymous.4open.science/r/panCancerDR-83EF/}) 
\end{abstract}

\section{Introduction}
To explore the drug responses of \textit{in vitro} cancer cells, several projects have utilized high-throughput profiling to assess cell viability when subjected to varying drug concentration treatments and yielded half-maximal inhibitory concentration. For instance, the Genomics of Drug Sensitivity in Cancer project (GDSC)~\cite{yang2012genomics} has assayed the sensitivity of more than one thousand of cancer cell lines to more than two hundreds of compounds. The Cancer Cell Line Encyclopedia (CCLE)~\cite{Barretina2012} is another effort that compiles genomic, transcriptomic, and drug sensitivity data for over 1,000 cancer cell lines. These public resources promote the development of machine learning methods for predicting drug response based on gene expression profiles~\cite{he2022context,chawla2022gene,ma2021few}. However, while some drugs exhibit promising sensitivity against tumor cells cultured under laboratory conditions, such observations offer limited guidance for clinical drug selection due to substantial discrepancy between \textit{in vitro} cellular context and \textit{in vivo} physiological environment. This disparity results in predictive methods that perform well on cell lines being less effective in predicting drug response in patients.


Domain generalization (DG) is proposed to construct models that perform well on unseen domains without access to their data during training. Recent advancements in DG have achieved remarkable progress~\cite{wang2022generalizing,zhou2021domain,carlucci2019domain}. However, few DG method has been proposed to clinical drug-response prediction tasks. On the other hand, Fourier transform has recently been utilized as a substitute for self-attention mechanism~\cite{lee2021fnet} or for time-series modeling \cite{koren2024interpretable,ye2023state}. Notably, FAN~\cite{dong2024fan} demonstrated that replacing conventional neural nodes with Fourier-transform operations not only significantly reduces the number of network parameters but also markedly enhances the model’s ability to fit periodic functions. Although a few studies have introduced Fourier transform into single-cell expression profiling \cite{nouri2025single,jia2023scgenerythm}, none explicitly formulate Fourier transforms and attention mechanisms within a unified framework for expression-based drug-response prediction.

In this paper, we proposed FourierDrug to generalize the predictive capacity of drug response on cell lines to out-of-distribution samples, such as individual cells and patients. We conceptualize each cancer type as a distinct source domain, with its cell lines serving as domain-specific samples, and then employed adversarial domain generalization to capture essential task-relevant features across multiple source domains. In particular, we are inspired by the observation that an anticancer drug exhibit initial effectiveness, but cancer cells often develop resistance through various biological mechanisms. The phenomenon suggests that drug-sensitive cells share common feature, whereas drug-resistant cells exhibit diverse and heterogeneous traits. So, we propose a Fourier asymmetric attention constraint that drives the sensitive samples aggregated into a single compact cluster, while resistant samples dispersed across frequency space. To validate its performance, we firstly evaluate it on bulk RNA-seq and drug response data from cell lines, using a leave-one-out validation strategy to assess its generalizability in predicting drug responses for unseen cancer types during training. The results demonstrated that our model achieved superior predictive performance across ten major cancer types. Moreover, we applied the model, trained exclusively on bulk RNA-seq of cell lines, to single-cell and patient-level prediction tasks. The results confirmed it achieved better or comparable performance compared to current state-of-the-art (SOTA) methods. Moreover, our model effectively captured the dynamic transition to resistance when cancer cells subjected to persistent drug exposure. These results highlight its potential as a ``train once, adapt anywhere" tool for advancing precision oncology.


\section{Related Works}
\subsection{Domain Generalization}
Domain generalization (DG) is proposed to construct models that perform well on unseen domains without access to their data during training. Recent advancements in DG have achieved remarkable progress~\cite{wang2022generalizing}, with approaches generally categorized into three main groups: data manipulation~\cite{shankar2018generalizing,yue2019domain,zhou2021domain}, representation learning~\cite{ghifary2015domain,li2018domain,shao2019multi,sicilia2023domain,zhang2022towards}, and learning strategy~\cite{chen2022discriminative,tian2022neuron,carlucci2019domain}. While our method falls within the realm of domain generalization, it significantly differs from existing domain adversarial learning methods. Specifically, we introduce a Fourier asymmetric attention constraint, enabling the model to better capture the nuances of real-world drug response data. The most closely related work is the SSDG\cite{jia2020single}, a single-sided domain generalization approach for face anti-spoofing. However, our method distinguishes itself through the incorporation of Fourier transform and asymmetric attention constraints, allowing the encoder to learn non-redundant features and thereby enhance the model's generalization capability.

\subsection{Fourier Transform in Deep Network}
Recent studies have increasingly introduced Fourier Transform (FT) into deep learning to enhance performance across diverse domains, including natural language processing (NLP) \cite{lee2021fnet}, time series modeling \cite{yi2023survey,ye2023state}, and the approximation of periodic functions \cite{dong2024fan}. For instance, FNet \cite{lee2021fnet} replaces the self-attention mechanism with Fourier transform, achieving 92\% of BERT's accuracy on the GLUE benchmark for natural language understanding tasks while delivering a 7-fold increase in training speed. For time series modeling, Neural Fourier Transform (NFT)~\cite{koren2024interpretable} integrates multidimensional Fourier Transform with temporal convolutional networks for multivariate time series forecasting, establishing new performance benchmarks across multiple datasets while also enhancing model interpretability. Notably, Fourier Analysis Network (FAN)~\cite{dong2024fan} integrates Fourier Transform operations into neural networks to improve their capacity for modeling periodic functions, achieving superior performance in real-world applications such as time-series forecasting and language modeling. While inspired by FNet~\cite{lee2021fnet}, our approach differs from it by introducing a novel constraint in the frequency domain analogous to the attention mechanism.


\subsection{Drug Response Prediction}
The prediction of clinical drug responses has drawn considerable attention from machine learning community. Some studies employed patient drug response data to fine-tune models initially trained on cell line datasets. CODE-AE~\cite{he2022context} utilizes domain separation network~\cite{bousmalis2016domain} to extract shared features between cell lines and patients, thereby predict the clinical drug response. Precily ~\cite{chawla2022gene} integrated signaling pathway and drug feature to predict drug responses \textit{in vitro} and \textit{in vivo}. In contrast, drug response prediction at the single-cell level is still in its infancy. Only a few studies leveraged domain adaptation to predict drug sensitivity of individual cells. For example, scDEAL ~\cite{chen2022deep} aligned the bulk RNA-seq and scRNA-seq features by minimizing the maximum mean discrepancy (MMD)~\cite{gretton2012kernel} in the latent space for single-cell drug response prediction. Distinct from the existing methods, we aim to establish a predictive model with superior generalizability across distinct target domains, spanning both single-cell and patient-level datasets, without access to target-domain data during training.

\section{Methods}
\subsection{Problem definition}
Assume we have $M$ domains $D=\{D_1,D_2...D_M\}$, with each domain corresponds to a specific cancer type. Each domain $D_k$ comprises the gene expression profiles of $N_k$ cell lines regarding to $k$-th cancer type, denoted as $D_k=\{X_1,X_2,\ldots X_{N_k}\}$. For a given drug, the response labels of the $N_k$ cell lines in domain $D_k$ are represented by $Y_{k}=\{y_{1},y_{2},\ldots,y_{N_k}\}$. Our primary objective is to build a deep learning-based model that accurately map the gene expression profiles of the cell line to their respective drug response labels. Once trained, the model can generalize effectively to predict drug responses in target domains, such as single-cell or patient samples.

\begin{figure*}[htpb]
    \centering
\centerline{\includegraphics[width=\linewidth]{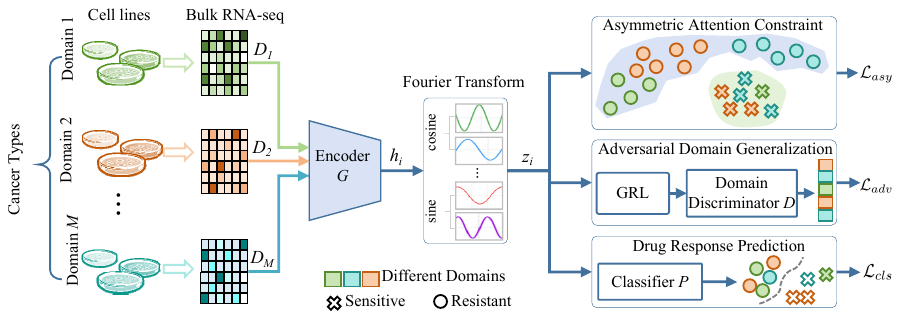}}
    \caption{Illustrative diagram of FourierDrug architecture characterized by Fourier asymmetric attention constraint (FAAC) module and asymmetric adaptive clustering constraint, with the adversarial domain generalization implemented by gradient reversal layer (GRL).} 
    \label{fig:framework}
\end{figure*}

\subsection{FourierDrug Framework}
The proposed architecture consists of five components: a shared encoder, a Fourier transform module, an asymmetric attention constraint, a classifier, a domain discriminator. As shown in Figure~\ref{fig:framework}, the expression profiles of cell lines are taken as input to the encoder for feature extraction. Next, Fourier transform is introduced to project the the encoded feature to frequency space, in which we impose an asymmetric attention constraint that would aggregate sensitive samples tightly together while dispersing resistant samples. Meanwhile, the domain discriminator aims to distinguish the domain of each sample (i.e., to identify the cancer type from the expression profile). The encoder and domain discriminator are adversarially trained so that the encoder is incentivized to learn features that make it increasingly difficult for the domain discriminator to correctly identify the domains. This adversarial training enables the encoder to capture domain-invariant features pertinent to drug response. Trained exclusively on bulk RNA-seq data from diverse cancer cell lines, FourierDrug achieves strong generalization to unseen pan-cancer types in cross-domain prediction tasks, including single-cell and patient-level drug response. 

\subsection{Fourier Asymmetric Attention Constraint}
Anticancer drugs are typically designed to target specific molecules or disrupt the biological processes critical for tumor cell proliferation. So, we observed that tumor cells are initially sensitive to some specific drugs, and subsequently develop resistance through diverse biological mechanisms, such as genetic mutations, alterations in cell cycle checkpoints, activation of alternative proliferation signaling pathways, and epigenetic modifications. Without loss of generality, we assume that sensitive cells share common pattern in the frequency domain, while resistant cells exhibit diverse heterogeneity. To reflect this observation, we introduce an Fourier asymmetric attention constraint (FAAC) into the domain generalization framework.

Assume that the feature space is $\mathbb{R}^{d}$, we construct a set of function bases isomorphic to it, mapping the feature space into frequency space. Formally, in the continuous domain over the interval $[-T, T]$, the sine and cosine functions form a complete orthogonal basis $\Phi = \{\phi_k(x), \psi_k(x)\}$, defined as:
\begin{equation}
\phi_k(x) = \cos\left(\frac{2\pi k x}{T}\right), \quad \psi_k(x) = \sin\left(\frac{2\pi k x}{T}\right), \quad k \in \mathbb{N}.
\end{equation}
A function $f(x)$ defined on $[-T, T]$ can be expanded using this basis as a Fourier series:
\begin{equation}
f(x) = \sum_{k=1}^{\infty} \left( a_k \phi_k(x) + b_k \psi_k(x) \right),
\end{equation}
where $a_k$ and $b_k$ are Fourier coefficients. Because the encoded features are represented as discrete vectors, we discretize the interval $[-T, T]$ into $d$ uniformly spaced sampling points $\{x_1, x_2, ..., x_d\}$. We then construct the discrete orthogonal basis vectors by sampling the sine and cosine functions at these points:
$$
\phi_k = \left[\cos\left(\frac{2\pi k x_1}{T}\right), ..., \cos\left(\frac{2\pi k x_d}{T}\right)\right]^\top, \quad
\psi_k = \left[\sin\left(\frac{2\pi k x_1}{T}\right), ..., \sin\left(\frac{2\pi k x_d}{T}\right)\right]^\top.
$$

For each feature vector $h_i \in \mathbb{R}^d$ extracted by encoder $G$, we project it onto the discrete orthogonal basis to obtain its representation in the frequency domain:
\begin{equation}
z_i\triangleq \mathcal{F}(h_i)=\sum_{k=1}^{d/2} \langle h_i, \phi_k \rangle \phi_k + \langle h_i, \psi_k \rangle \psi_k,
\end{equation}
where $\langle \cdot, \cdot \rangle$ denotes the Euclidean inner product. Note that $z_i$ does not represent the Fourier expansion, it instead corresponds to a set of Fourier coefficients obtained from the Fourier expansion. As a result, the transformation maps the latent representation of expression profiles into frequency domain, enabling explicit control over frequency spectrum. To enhance the alignment of semantically similar samples in the frequency domain, we introduce an asymmetric  constraint based on the Fourier-transformed representations. Specifically, given a positive sample $z_i$ as an anchor, we aim to maximize the cosine similarity between the anchor and positive samples, while minimizing the cosine similarity between the anchor and negative samples. Importantly, we do \emph{not} impose any constraint between the negative samples themselves. This leads to the following asymmetric loss function:
\begin{equation}
\mathcal{L}_{\text{asy}} = -\frac{1}{|\mathcal{P}_i|} \sum_{j \in \mathcal{P}_i} \frac{z_i^\top z_j}{\|z_i\| \cdot \|z_j\|} 
+ \frac{1}{|\mathcal{N}_i|} \sum_{k \in \mathcal{N}_i} \frac{z_i^\top z_k}{\|z_i\| \cdot \|z_k\|},
\end{equation}
where \( \mathcal{P}_i \) represents the set of positive samples (drug-sensitive), and \( \mathcal{N}_i \) represents the set of negative samples (drug-resistant). 


We further illustrate that the asymmetric constraint mathematically correlates to self-attention mechanism. In standard Transformer architectures~\cite{vaswani2017attention}, self-attention weight is computed via scaled dot-product similarity between the query and key embedding:
\begin{equation}
\alpha_{i,j} = \frac{q_i^\top k_j}{\sqrt{d_k}}.
\end{equation}

Since $z_i^\top z_j = \|z_i\| \cdot \|z_j\| \cdot \cos(z_i, z_j)$,
we can interpret the cosine similarity in frequency domain as a \textbf{normalized form} of the attention dot-product, thereby capturing angular correlations between frequency components of the input features. Furthermore, according to the Convolution Theorem, i.e. the element-wise multiplication in the frequency domain corresponds to convolution in the time domain, we can interpret \( z_i^\top z_j \) as an approximation of the operation between \( x_i \) and \( x_j \) through continuous-space convolution. Unlike the standard self-attention mechanism, our approach achieves global attention associations without increasing the number of parameters. Our experiments demonstrate that this approach not only enhances the model's generalizability but also effectively mitigates overfitting (See Section~\ref{sec:results}).


\subsection{Adversarial Domain Generalization}
The objective of domain generalization is to learn domain-invariant features relevant to the prediction task, while eliminating domain-specific information. For this purpose, we introduce a domain discriminator designed to classify the domain of each input sample. The encoder and domain discriminator are trained in an adversarial manner, with the encoder learning to extract features that prevent the discriminator from accurately identifying the domain. This adversarial training process encourages the encoder to capture features that are both predictive of drug response and consistent across all domains. Denote by $L_{adv}$ the cross entropy loss for domain discrimination, we have
\begin{equation}
   \min_D\max_G \mathcal{L}_{adv}=-\sum_{k=1}^{M}\sum_{i=1}^{N_k}(y_i^{(D)}\log D(z_{i})+(1-y_i^{(D)})\log(1-D(z_{i})))
\end{equation}
in which $y_i^{(D)}$ is the true domain label of the input sample $X_i$, $D(z_i)$ represents the domain label predicted by the domain discriminator. The domain discriminator $D$ aims to minimize the loss, whereas the encoder strives to maximize it. In our practice, we employ the Gradient Reversal Layer (GRL) to implement the adversarial training between the encoder and the domain discriminator. 
 
\subsection{Drug Response Prediction}
The drug response labels of cell lines subjected to specific drug exposure is used to train the classifier. It takes as input the Fourier-transformed features to predict the response labels. We use cross-entropy as the classification loss, with the loss function $\mathcal{L}_{cls}$ defined as: 
\begin{equation}
    \min_P\mathcal{L}_{cls}=-\sum_{k=1}^{M}\sum_{i=1}^{N_k}y_i\log(P(z_i))+(1-y_i)\log(1-P(z_i))
\end{equation}
in which $P(z_i)$ represents the predicted drug response label by the predictor. Finally, the full objective is defined as below:
\begin{equation}
\mathcal{L}=\mathcal{L}_{adv}+\lambda_1\mathcal{L}_{asy}+\lambda_2\mathcal{L}_{cls}
\end{equation}
where $\lambda_1$ and $\lambda_1$ are the balanced parameters. In our practice, the encoder are realized using fully-connected feed-forward networks with rectified linear unit (ReLU) activation function. It consist of only two feed-forward layers with sizes of 1024 and 740, respectively. Each feed-forward layer is followed by a batch normalization layer, and a dropout layer with the dropout probability set to 0.1. The learning rate is set to 8e-5. Our model was implemented in PyTorch 3.10, and all experiments were conducted on a CentOS Linux 8.2.2004 (Core) system, equipped with a GeForce RTX 4090 GPU and 128GB memory. During the model training and cross-validation stage, these loss terms were appropriately weighted.

\section{Results} \label{sec:results}
\subsection{Data Resource and Preprocessing}
We regard the bulk RNA-seq data from different cell lines of same cancer type as a source domain, with drug response (sensitive vs resistant) as class labels. The dataset were obtained from the Genomics of Drug Sensitivity in Cancer (GDSC) project~\cite{yang2012genomics}, which contained a wealth of data about the responses of 1074 cancer cell lines to 226 therapeutic agents. These cell lines come from more than 20 cancer types. The drug sensitivity were quantified using the half maximal inhibitory concentration (IC50). From the GDSC dataset, we first identified all cell lines treated by the drug of interest. Next, the cell lines were then ranked based on their IC50 values and categorized into two classes: sensitive (labeled as 1) and resistant (labeled as 0), with the threshold defined as the average IC50 value. Besides, because drug-induced changes in gene expression reflect the substantial effect of a drug on cellular phenotype, we selected 3,000 highly variable genes from the expression profiles of over 10,000 genes available in the bulk RNA-seq data of cell lines. These selected genes were used as inputs to our model.

We validated our model's generalizability not only on hold-out bulk drug response datasets, but also across multiple independent test sets. These test sets include single-cell drug responses of multiple types of cell lines to distinct drugs, and clinical drug response of patients from the TCGA cohort. In particular, we tested it a scRNA-seq dataset tracking the dynamic state transitions of breast cancer cells under persistent drug exposure. The detailed descriptions of these test sets are presented in respective experiments.

\subsection{Evaluation on Leave-Out Validation}
We first evaluated the performance of FourierDrug in predicting drug response for unseen cancer types during training. For objective evaluation, all cell lines of a specific cancer type were designated as the test set, while cell lines from other cancer types were used for training. This leave-one-out approach enabled us to evaluate the performance of FourierDrug on unseen cancer types for a specific drug. Notably, some cancer types have an insufficient number of samples (cell lines) for reliable performance evaluation. These cancer types were excluded from testing but were still included as source domains in the training set. As a result, the leave-one-out test set included ten cancer types. These drugs represent a diverse array of therapeutic classes, including chemotherapy agents, targeted therapies, and broad-spectrum inhibitors. 

\begin{figure*}[htb]
    \centering
\centerline{\includegraphics[width=\linewidth]{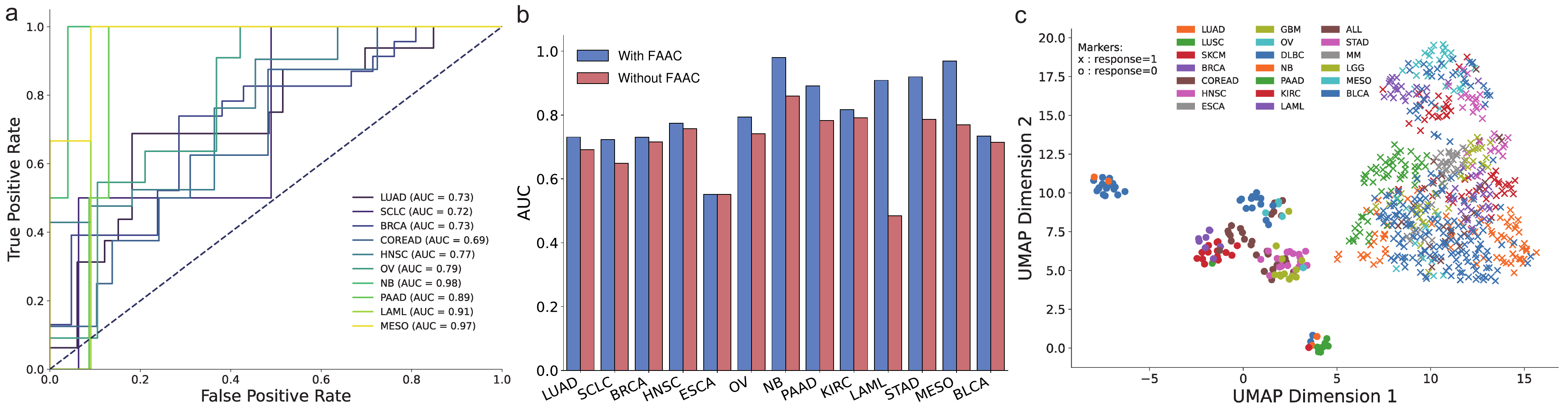}}
    \caption{Performance evaluation of FourierDrug in predicting bulk drug responses for hold-out cancer types. (a) ROC curves for ten hold-out cancer types. (b) AUROC values achieved by our model with and without Fourier asymmetric attention constraint (FAAC) module. (c) UMAP plots of the learned features of 20 cancer types included in the training set. } 
    \label{fig:bulk}
\end{figure*}

The experimental results demonstrated that our model achieved superior performance in predicting drug responses across ten major cancer types (Figure~\ref{fig:bulk}a), with particularly high AUC values exceeding 0.9 in the cancers such as neuroblastoma (NB), mesothelioma (MESO), and acute myeloid leukemia (LAML). Next, we validated the effectiveness of the FAAC module in improving performance. As illustrated in Figure~\ref{fig:bulk}b, inclusion of this module led to remarkably improved model performance. To further validate that the encoder captured the discriminative feature related to drug response, we used UMAP to project the learned features into the two-dimensional space. It can be found that the sensitive and resistant sample were clearly separated (Figure~\ref{fig:bulk}c). Particularly, the sensitive samples gathered into a cluster closely, whereas the resistant samples dispersed into multiple separate clusters. This results strongly verified the validity of the FAAC module. Moreover, we performed regression analysis between the learned features and the real IC50 values, comparing them to PCA features that accounted for 95\% variance across all genes. The results indicated that the features extracted by the encoder provided a better fit to the IC50 values. These findings confirmed that our model successfully captured drug response-related features from gene expression profiles, rather than simply memorizing drug response labels, thereby enabling it to predict the drug responses for unseen cancer types during training.

\subsection{Generalization to Single-Cell Domain}
The inherent heterogeneity of tumors often leads to significant variability in gene expression profiles of individual cancer cells within a tumor. Meanwhile, the noise present in single-cell RNA sequencing (scRNA-seq) data further complicates this issue, leading to data distributions that differ substantially from the bulk RNA-seq data used during model training. To assess the generalizability of our proposed method, we systematically evaluated its performance in predicting drug responses at the single-cell level.

\begin{figure*}[htbp]
    \centering
\centerline{\includegraphics[width=\linewidth]{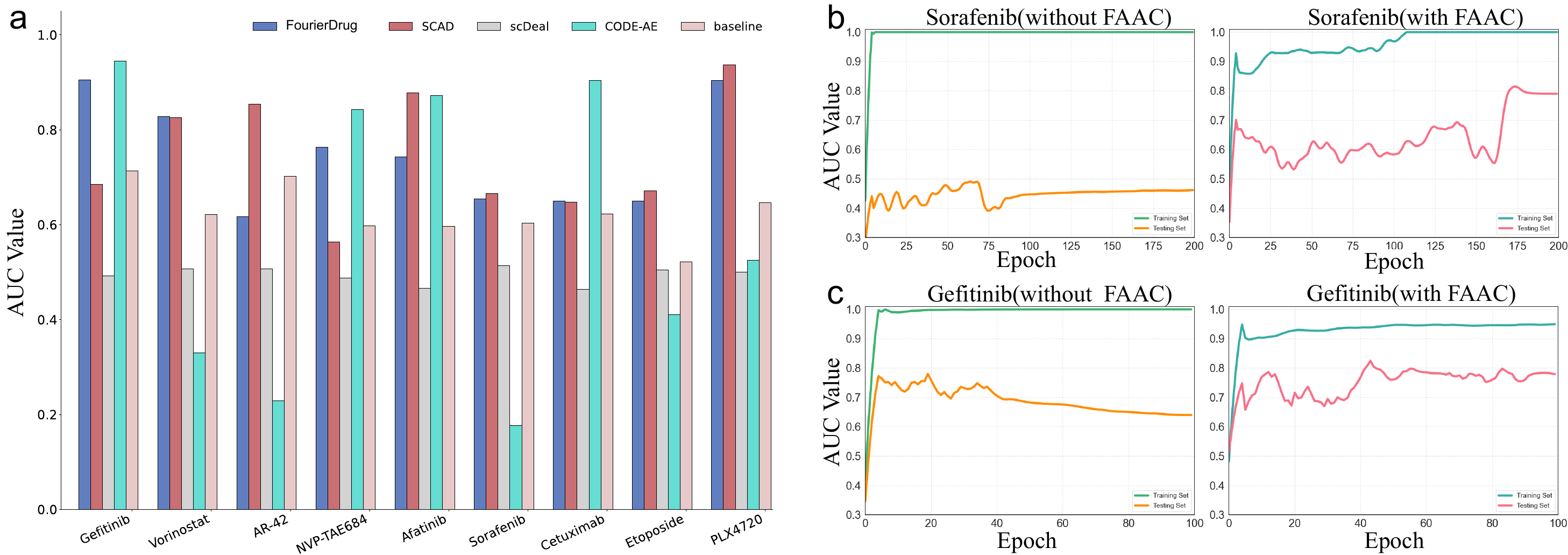}}
    \caption{Performance evaluation of FourierDrug trained on bulk RNA-seq in predicting single-cell drug response. (a) Performance comparison with baseline and SCAD for nine distinct drugs. (b-c) AUC curves with respect to training epochs of FourierDrug with and without the Fourier asymmetric attention constraint (FAAC), respectively.} 
    \label{fig:scRNA}
\end{figure*}

The single-cell drug response datasets used for performance evaluation comprised both pre-treatment scRNA-seq data from CCLE~\cite{Barretina2012} and post-treatment scRNA-seq data from GEO repository (accession numbers: GSE149215 and GSE108383). The pre-treatment dataset included the expression profiles and drug response labels of JUH006 cell line to the treatment of three distinct drugs (Gefitinib, Vorinostat and AR-42), as well as the data of SCC47 cell line treated with other four distinct drugs (NVP-TAE684, Afatinib, Sorafenib, Cetuximab). The post-treatment datasets contained the expression profiles and drug response labels of PC9 cell line following Etoposide treatment~\cite{aissa2021single}, as well as the A375 and 451Lu cell line treated with PLX4720~\cite{ho2018single}. For performance evaluation on scRNA-seq data, we selected a subset of highly variable genes that exhibited the most significant differences in expression levels across both bulk and single-cell RNA-seq data, and used these genes as inputs into our model. The details about the single-cell drug response datasets are listed in Table~S1.

To benchmark performance, we build a baseline model composed of only three fully-connected layers. The baseline model was trained on the GDSC dataset and directly applied to the single-cell datasets. Meanwhile, we conducted performance comparison to three state-of-the-art methods: SCAD~\cite{zheng2023enabling}, scDEAL~\cite{chen2022deep} and CODE-AE~\cite{he2022context}. SCAD and scDEAL are domain adaptation-based methods designed for predicting single-cell drug responses, while CODE-AE leverages feature disentanglement to extract common feature between source and target domains. We obtained the source codes of these competing methods, trained them using the GDSC dataset, and evaluated their performance on the single-cell datasets. The experimental results showed that our method remarkably outperformed the baseline model across all drugs (Figure~\ref{fig:scRNA}a). Particularly, compared to SCAD that used SMOTE sampling for class balance and top 4k highly variable gene as input (smote\_tp4k), our method achieved better or comparable performance for most drugs. The performance of CODE-AE is highly unstable. While it achieves outstanding results for specific drugs, such as Gefitinib and Cetuximab, it performs poorly for other drugs, including Vorinostat, AR-42, Sorafenib, and Etoposide. Notably, these competing methods utilized the scRNA-seq data as target domain during the training stage, while our method never used scRNA-seq data during training. This highlights the ability of our method to extract generalizable features related to drug response from gene expression profiles across diverse source domains via domain generalization. 

Furthermore, we assess the effectiveness of the FAAC module in improving model generalizability. By comparing the performance with and without this module, we evaluate its impact on predicting single-cell drug response. The experimental results demonstrate that incorporating the FAAC module effectively prevents overfitting (Figure~\ref{fig:scRNA}b, Figure S1), thereby improving our model's generalizability to out-of-distribution data.


\subsection{Generalization to Patient-Level Domain}
For further evaluation, we applied FourierDrug to predict clinical drug responses in patients. \textit{In vivo} drug response prediction poses significant challenges due to the influence of many biochemical factors, making it inherently more difficult than \textit{in vitro} predictions on cell lines. This experimental setting provides a more stringent evaluation of the model’s generalizability. 

The expression profiles and clinical metadata of patients were obtained from the TCGA repository~\cite{hutter2018cancer}. The patients treated with one of four drugs—Fluorouracil, Gemcitabine, Temozolomide, or Cisplatin—were selected for analysis. The four drugs were chosen due to the relatively large number of patients subjected to the treatments of these drugs, enabling objective and reliable performance evaluations. Patients exhibiting a complete response or partial response were labeled as sensitive, while those with clinically progressive or stable disease were labeled as resistant. From the patient expression profiles, 3,000 differentially expressed genes (DEGs) were identified based on a significance threshold of p-value$\leq$0.05. The DEGs were then intersected with gene expression data from the GDSC dataset, and the overlapping genes were subsequently used as inputs for our model.

\begin{figure}[htbp]
    \centering
    \includegraphics[width=\linewidth]{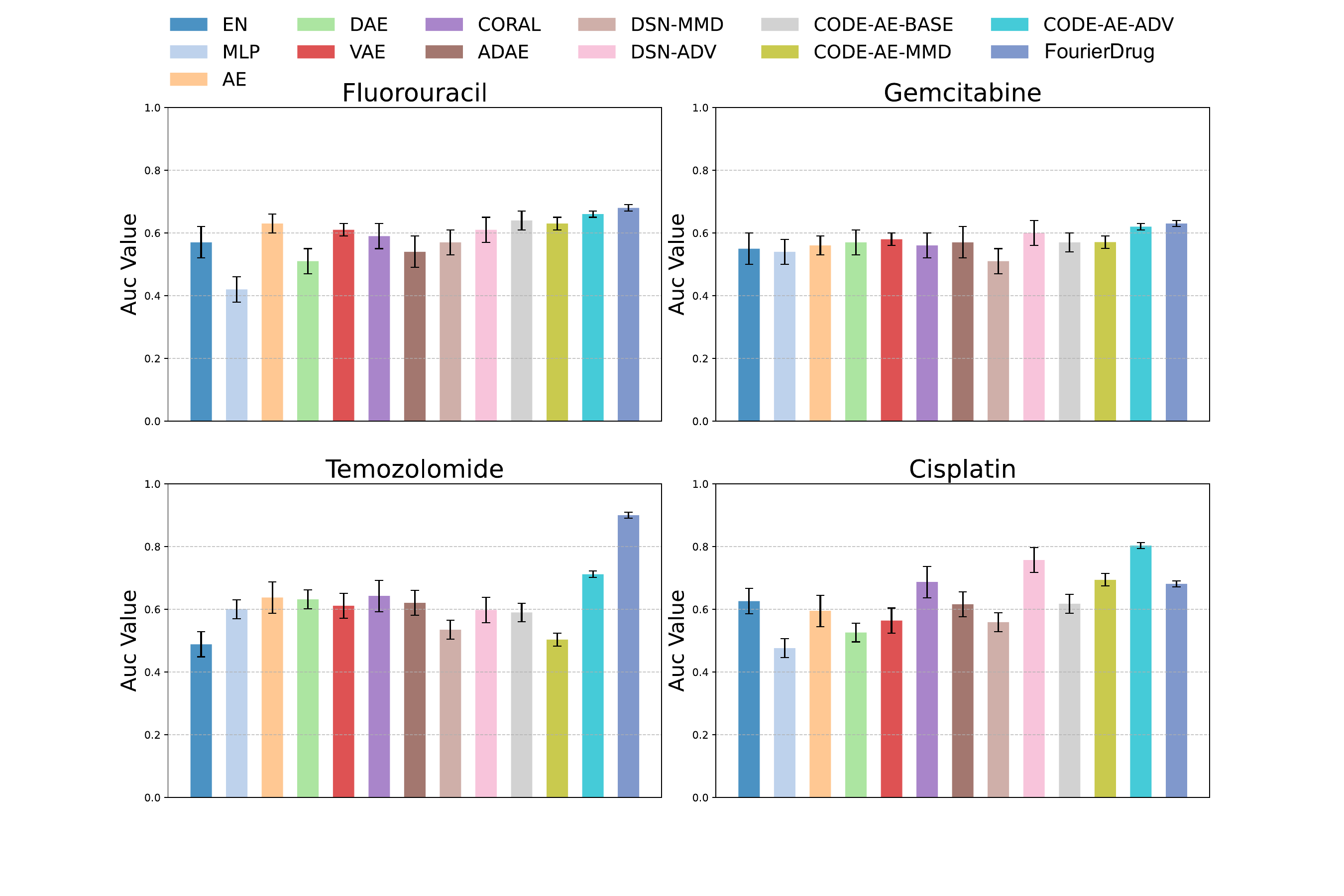}
    \caption{Performance comparison of FourierDrug with ten existing methods on TCGA patient drug response prediction.}  \label{fig:patient}
\end{figure} 

For performance evaluation, we compared it with ten previously published methods. These methods include conventional machine learning classifiers (e.g., MLP, EN) and deep learning models (e.g., AE, VAE, DAE), as well as various domain adaptation methods (e.g., ADAE, CORAL, DSN variants, CODE-AE variants). As illustrated in Figure~\ref{fig:patient}, our model demonstrated superior performance across all drugs except Cisplatin, where it marginally underperformed relative to CODE-AE-ADV~\cite{he2022context} and DSN-ADV~\cite{bousmalis2016domain} but outperformed other methods. Notably, for Temozolomide, our method remarkably outperformed all competing methods. 

Furthermore, we conducted ablation studies on four additional drugs—Docetaxel, Paclitaxel, Sorafenib, and Vinorelbine—to evaluate the contribution of the latent independent projection module. The results verified that the inclusion of this module brought substantial performance gains, exceeding 10\% across all four drugs (Figure S2). The most pronounced improvement was observed for Sorafenib, with an AUC increase by over 30\%.

\subsection{Evaluation on Dynamic Drug Response Transition}
\begin{figure*}[htbp]
    \centering
    \includegraphics[width=\linewidth]{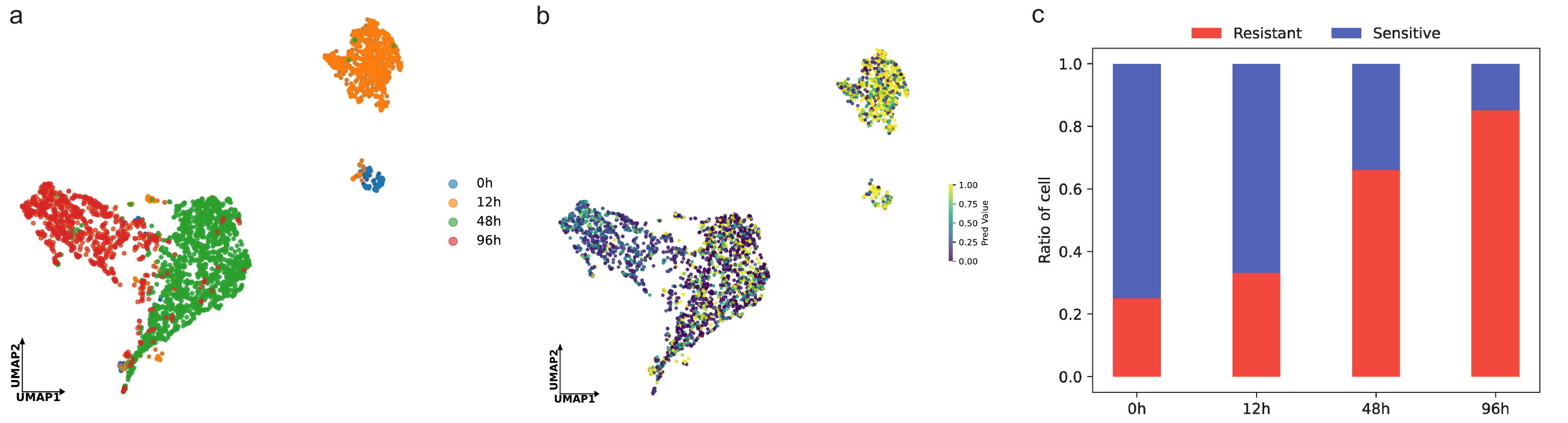}
    \caption{Accurate prediction of dynamic transition to resistance of MCF7 cells in response to Bortezomib treatment at four time points. (a-b) UMAP plots of the scRNA-seq data all cells exposed to Bortezomib colored by time points and predicted sensitivity scores, respectively. (c) Percentage of sensitive and resistant cells predicted by FourierDrug at each time point. }  \label{fig:series}
\end{figure*}

Cancer cells are known to rapidly evolve under therapeutic pressure, often developing resistance to specific drugs. To evaluate the generalizability of FourierDrug, we investigated its ability to predict temporally dynamic changes in individual cell states during prolonged chemotherapy exposure. For this analysis, we utilized the dataset published by Ben-David et al.\cite{ben2018genetic}, comprising 7,440 single-cell clones derived from the MCF7 cell line. These clones were exposed to 500 nM Bortezomib for two days and underwent single-cell sequencing at distinct time points: prior to treatment (t0, $n=160$), 12 hours post-treatment initiation (t12, $n=994$), 48 hours post-exposure (t48, $n=1,623$), and 96 hours following drug washout and recovery (t96, $n=963$). UMAP visualization revealed four transcriptionally distinct clusters corresponding to the time points analyzed (Figure\ref{fig:series}a), indicating significant transcriptional divergence driven by drug exposure.

To assess whether FourierDrug could accurately capture the state transitions of MCF7 cells under sustained Bortezomib exposure, we applied it to predict drug response states and analyzed the predicted sensitivity scores of individual cells (Figure~\ref{fig:series}b). The predictions showed that nearly all MCF7 cells were highly sensitive to Bortezomib at baseline (t0). However, with increasing exposure duration, the proportion of sensitive cells gradually declined, accompanied by a rise in resistant cell populations. By 96 hours post-treatment (t96), the cell population exhibited the highest fraction of resistant cells (Figure~\ref{fig:series}c). These results closely aligned with observations from prior studies~\cite{pellecchia2023predicting}, further substantiating the model's predictive accuracy. Our findings demonstrate that FourierDrug effectively captures the temporal dynamics of cellular responses to drug treatment across multiple time points, highlighting its potential as a robust tool for modeling drug-induced phenotypic evolution in cancer cells.

\section{Discussion and Conclusion}
In summary, we proposed a novel domain generalization method for predicting drug responses without needing target-domain data during training. By employing Fourier transforms and asymmetric attention constraints, the approach aggregates sensitive samples and disperses resistant ones, achieving robust feature extraction. Validated on bulk RNA-seq, single-cell, and patient-level data, FourierDrug outperformed state-of-the-art methods, demonstrating exceptional cross-domain generalizability. Additionally, its capability to predict dynamic resistance transitions underscores its clinical relevance. This framework bridges preclinical and clinical applications, providing a scalable ``train once, adapt anywhere" solution for advancing precision oncology.

\bibliographystyle{unsrt}
\bibliography{reference}

\end{document}